\documentclass[conference]{IEEEtran}
\usepackage{graphicx} 
\usepackage{subfig}
\usepackage[hyphens]{url}

\begin{document}
\title{TotalBotWar: A New Pseudo Real-time Multi-action Game Challenge and Competition for AI}

\author{\IEEEauthorblockN{Alejandro Estaben, C\'{e}sar D\'{i}az, Raul Montoliu}
\IEEEauthorblockA{Insitute of new imaging Technologies\\
Jaume I University\\
Castell\'{o}n, Spain\\
Email: montoliu@uji.es}
\and
\IEEEauthorblockN{Diego P\'{e}rez-Liebana}
\IEEEauthorblockA{Game AI group\\
Queen Mary Universtiy of London\\
London, UK \\
Email: diego.perez@qmul.ac.uk}
}

\maketitle

\begin{abstract}
This paper presents \textit{TotalBotWar}, a new pseudo real-time multi-action challenge for game AI, as well as some initial experiments that benchmark the framework with different agents. The game is based on the real-time battles of the popular TotalWar games series where players manage an army to defeat the opponent’s one. In the proposed game, a turn consists of a set of orders to control the units. The number and specific orders that can be performed in a turn vary during the progression of the game. One interesting feature of the game is that if a particular unit does not receive an order in a turn, it will continue performing the action specified in a previous turn. The turn-wise branching factor becomes overwhelming for traditional algorithms and the partial observability of the game state makes the proposed game an interesting platform to test modern AI algorithms.
\end{abstract}

\IEEEpeerreviewmaketitle

\section{Introduction}
In recent years, games have proven to be important test-beds for Artificial Intelligence (AI). For instance, deep reinforcement learning has enabled computers to learn how to play games such as Chess \cite{Silver18}, Go \cite{Silver18}, Atari games \cite{Mnih15}, and many other games \cite{Justesen17}. Despite these important advances, there are still games that pose important challenges for state-of-the-art AI agents. Some examples are \textit{Blood Bowl} \cite{Justesen19}, \textit{Legend of Code and Magic} \cite{LOCM19}, \textit{MicroRTS} \cite{Ontanon18}, \textit{FightingICE} \cite{Ishii19}, \textit{Hanabi} \cite{walton2019}, \textit{Splendor} \cite{bravi2019}, \textit{StarCraft} \cite{starCraftAICompetition}, and the \textit{General Video Game AI} framework \cite{gvgaibook2019}, among others. 

In this paper, we propose \textit{TotalBotWar}, a new pseudo real-time challenge for game AI. The game is inspired by the real-time battles of the popular \textit{TotalWar} game series\footnote{Creative Assembly, \url{https://www.totalwar.com/}}, where two players control respective armies with the objective of defeating each other. On each turn, the agent must decide where the unit must move to. When two opposite units collide, they will start to fight. The result of the combat depends on the type of units and their attributes. If during a turn a unit does not receive any order, it will continue its movement following the previous one, or it will stand still if none was given. This introduces unknown information on the state: it is possible to know that an enemy unit is moving, but not its destination. The game has a high number of possible actions in a turn ($\approx6.7E7$) and also a huge number of possible states ($\approx3.3E29$), which provides a significant challenge for AI agents. The game, implemented using the \textit{CodinGame} SDK\footnote{https://www.codingame.com/}, has already been made available online at this platform\footnote{https://www.codingame.com/contribute/view/486222077fe22e3aa6bcdc0f729dd46223bb}.

An initial set of experiments are also presented, where five different agents are benchmarked to give a baseline to future researchers. Three of them are primary agents where a) units never move (but can fight), b) always move forward, or c) move to a random localisation. The two remaining are more sophisticated. The first one applies human knowledge by using a heuristic function. The last one implements the Online Evolutionary planning (OEP) algorithm proposed by Justesen et al. \cite{Justensen17journal}. Preliminary results show that the heuristic-based and OEP overcome the three primary agents, being the OEP preferable.

Summarising, this paper presents three main novelties:

\begin{itemize}
    \item We present \textit{TotalBotWar} a new pseudo-real-time multi-action challenge for game AI.
    \item As far as we know, this is the first work implementing real-time \textit{TotalWar}-style battles as a game AI challenge.
    \item We assess the performance of five agents, including the Online Evolutionary Planning algorithm, in the proposed game AI challenge. 
\end{itemize}

The rest of the paper has been organised as follows. Section \ref{sec:game} presents the main characteristics of the game including how agents interact with the game engine. A set of baseline agents and some preliminary experiments are showed in Sections \ref{sec:agents} and \ref{sec:exps}. Finally, the most important conclusions drawn from this work are summarised in Section \ref{sec:conclusions}.

\section{TotalBotWar}
\label{sec:game}

\subsection{Game Overview}
\emph{TotalBotWar} is a 1 vs 1, pseudo real-time, multi-action game, partially inspired in the real-time battles of the \emph{Total War} games series. In our game, both players start with the same number of military units and their objective is to defeat the other player. The winner is the player who first destroys all the opponent’s units or the one with more units alive on the battlefield when the maximum number of turns is reached, which is set to $400$.

\begin{figure*} [!t]
\centering
\subfloat{\includegraphics[width=0.12\textwidth]{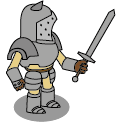}}  
\subfloat{\includegraphics[width=0.12\textwidth]{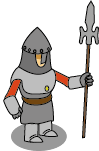}}  
\subfloat{\includegraphics[width=0.12\textwidth]{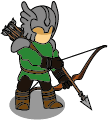}}  
\subfloat{\includegraphics[width=0.12\textwidth]{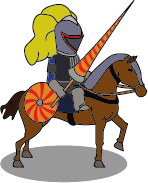}}  
\caption{The four game units, from left to right: swordsmen, spearmen, archers and knights.}
\label{fig:units}
\end{figure*}

There are four different types of units: Swordsmen, Spearmen, Archers and Knights (see Figure~\ref{fig:units}). The game uses a classical \emph{rock-paper-scissors} combat scheme, where swordsmen are good against spearmen, spearmen are good versus knights and finally, knights are good against swordsmen. Archers are an exception: they can attack from distance but are very weak in a face to face combat. 

Each unit has an attribute vector modelling its behaviour. The attributes are Health Points, Attacking Strength, Defence, Charge Power, Charge Resistance, Moving Speed and defence against Arrows. Besides, archers also have Throwing Distance and Arrow Damage. Table \ref{tab:attributes} shows the values assigned to each attribute for each unit type. 

\begin{table*}[t]
\centering
\caption{Values of the attributes of each unit type.}
    \begin{tabular}{|c|c|c|c|c|}
    \hline
    Attribute & Swordsmen & spearmen & Knights & Archers \\
    \hline
    \hline
    Health Points          & 250 & 250 & 200 & 100 \\
    Attacking Strength     &  20 &  15 &  12 &  10 \\
    Defence                &  10 &  20 &  12 &   5 \\
    Charge Power           &   5 &  10 & 100 &   5 \\
    Charge resistance      &  25 & 125 &  15 &   0 \\
    Moving Speed           &  15 &  10 &  40 &  15 \\    
    Defence against Arrows &  10 &  30 &  30 &  10 \\
    Throwing Distance      &  -  & -  & -    & 450 \\
    Arrow damage           &  -  & -  & -    & 20 \\
    \hline
    \end{tabular}
\label{tab:attributes}
\end{table*}

Units can move to any place of the battlefield. Two units from the same agent can't overlap, while they will fight if belonging to different armies. If an unit reaches the limits of the battlefield, it stops. Archers always shoot arrows to enemy troops into the attacking range. Troops suffer friendly-fire if they are close to an opponent unit receiving arrows.

The game has three different leagues or levels. When using the \textit{CodinGame} platform, the player has to first implement a bot to defeat the system bot of the first league. After, he/she must implement a new one to defeat the system bot of the second league before passing to the third one. In the third league, the \textit{CodingGame} platform allows testing the player's bot versus the bots implemented by many other players. Alternatively, the three leagues can be used isolated from \textit{CodingGame} platform to test AI algorithms and to rule AI competitions. We plan to use the third league as a future Game AI competition.

\begin{figure}[!t]
\centering
\includegraphics[width=0.99\linewidth]{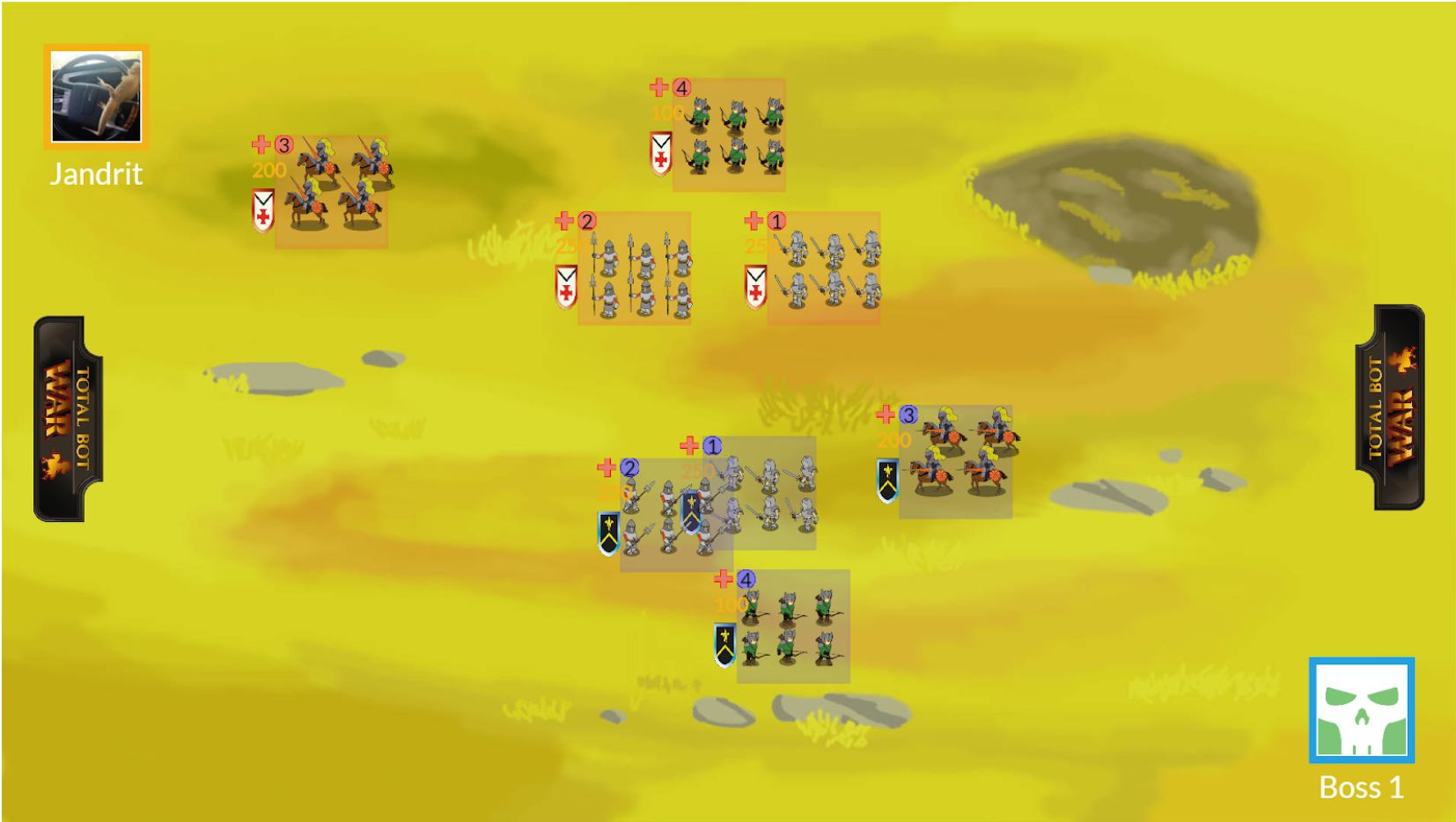}
\caption{Screenshot of the first league of the game. The army has 4 units, one of each type.}
\label{fig:liga1}
\end{figure}

\begin{figure}[!t]
\centering
\includegraphics[width=0.99\linewidth]{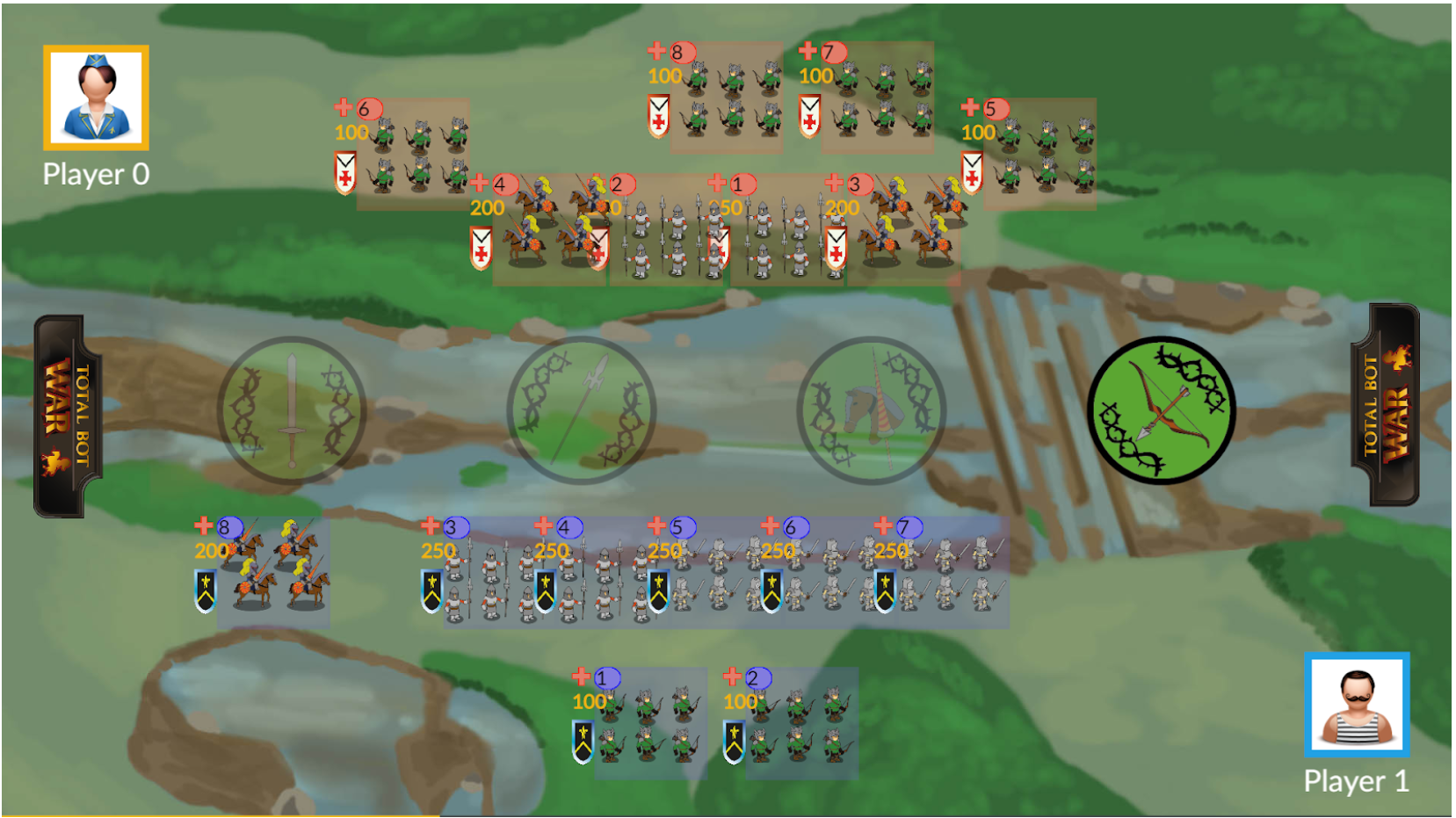}
\caption{Screenshot of the second league of the game during the draft phase. In this league, the army is composed by 9 different units. The composition of the army is defined at draft phase. }
\label{fig:liga2}
\end{figure}

\begin{figure}[!t]
\centering
\includegraphics[width=0.99\linewidth]{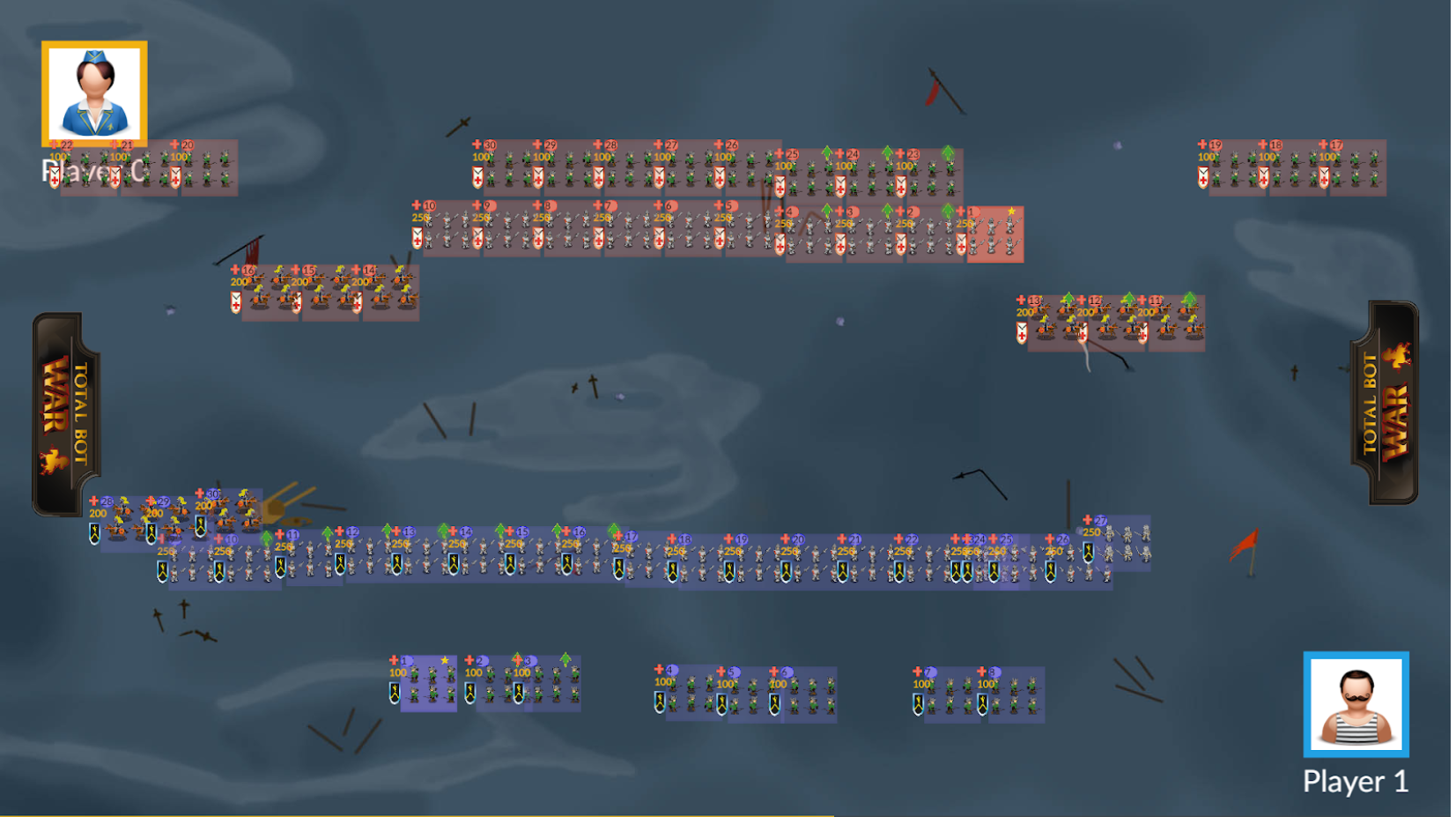}
\caption{Screenshot of the third league of the game during the battle phase. The army is composed by 30 units and includes a General. The General is the unit with ID 1 and its background colour is highlighted.}
\label{fig:liga3}
\end{figure}

In the first league (see Figure \ref{fig:liga1}), the army of each player consists of just one unit of each type and units start in predefined initial locations on the battlefield. The second league introduces the draft phase (see Figure \ref{fig:liga2}) where, in the first 9 turns of the game, the agents must select how many units of each type will be part of their armies and their initial positions on the battlefield. Therefore, the total number of units is 9. There is no restriction on the number of units of each type, i.e. the army can be composed by 9 archers if this is the decision of the agent. On each turn, players select units simultaneously, knowing only the units selected by both players in all previous turns.

In the third league (see Figure \ref{fig:liga3}), the army is composed of 30 units, therefore the draft lasts until the $30^{th}$ turn of the game. Additionally, the third league introduces the \textit{General} unit, which is a highly important unit that can be crucial in the game. All units within a distance of $150$ pixels to their general increase (multiplicatively) their attributes by $1.25$. However, if the general is dead, all units decrease their attributes by $0.75$. The general is always the first unit selected in the draft and can be of any type. 

The size of the battlefield is $1920\times1080$ pixels. The size of the units is $150\times150$ pixels in leagues 1 and 2, and $75\times75$ pixels in the third league. 

\subsection{Main characteristics}
The main characteristics for game AI are as follows:
\begin{itemize}
\item It is a 1 vs 1 game.
\item It is (pseudo) real-time. Although the game engine performs actions in the order indicated in the turn, the effect of this order is practically negligible. Similarly, the effect of which player performs first the actions is minimal.
\item Not all information is known in the state. The state contains information about the actual position of the enemy units and if they are moving or not, but it does not provide information about the final target where they are moving.
\item It is multi-action since in the same turn more than one action can be performed for each different unit owned in the battlefield.
\item The agents have just $200$ms to decide the actions to be executed on each the turn. This is a restriction of the core of \textit{CodingGame} engine.
\item It has a very large number of possible actions in a turn ($\approx6.7E7$) and possible states ($\approx3.3E29$).
\end{itemize}

\subsection{Motivation}
This work has two principal motivations. On the one hand, the game has been included in \textit{CodinGame} platform to be used as a tool to learn programming skills fascinatingly.  \textit{CodinGame} platform allows the use of many programming languages, and it is possible to see the effect of the source code used for the agent developed. That can help beginners to learn programming languages faster than through a more traditional style of teaching. The first league of the game is perfect for this purpose. On the other hand, the second and third leagues are dedicated to the developing of new game AI agents due to their complexity. The third league is the one selected to be a game AI competition in the future.

The develop a new game AI challenge using the \textit{CodinGame} SDK has three main advantages with respect to completely develop it from zero:

\begin{enumerate}
    \item  Developers can take advantage of the framework with contains useful code that can make easier to develop a new game. 2
    \item Users can program their agents in their preferred programming language instead of being restricted, as usually happens, to use just the one used to develop the game. 
    \item Sometimes to start with a new game AI challenge is hard since, for instance, users not always have installed the correct libraries to run the game. The use of \textit{CodinGame} platform avoid this kind of problems.
\end{enumerate}

However, some constraints must be accomplished as the maximum size of the battlefield, the amount of \textit{thinking} time per turn for the agents and the maximum number of turns, among others. 

The game is inspired in \textit{Total War} games since they are very popular for the general public. Similarly to other popular games AI challenges as \textit{StarCraft} AI competition, this can engage students to learn programming languages in general and AI in particular since they can be highly motivated to develop agents to play popular games. 

\subsection{Action Space}
On each turn, the current player can provide an action for each one of their units. An action consists of moving a unit a particular number of pixels in both $x$ and $y$ directions and the movement normally takes several turns to be completed. If in a turn the player does not indicate an action for a particular unit, it continues the movement following the previous action performed on this unit.

An action has the following format: ``\emph{ID} $\delta_x$ $\delta_y$'' where:

\begin{itemize}
\item \emph{ID} is the unique ID of the unit.
\item $\delta_x$ is the number of pixels we want to move the troop on the X axis.
\item $\delta_y$ is the number of pixels we want to move the troop on the Y axis.
\end{itemize}

Note that $\delta_x$ and $\delta_y$ are not the global coordinates to move the unit to, but how many pixels the unit must move with respect to its current coordinates. The coordinates are relative to the unit location to be independent to the position of the agent (up or down) in the battlefield.

For instance, some actions that can be played are:

\begin{itemize}
\item ``1 100 50'': Unit with ID $1$ will move $100$ pixels to its right (east in the display if the agent plays in the bottom part of the battlefield or west, otherwise) and $50$ to the front of the battlefield (upwards or north in the display if the agent plays in the bottom part of the battlefield or downwards or south, otherwise).
\item ``3 -100 -10'': Unit with ID $3$ will move $100$ pixels to its left and $10$ pixels backward.
\item ``5  0  0'': Unit with ID $5$ will stop.
\end{itemize}

On each turn, a player can perform more than one action and they are indicated as a string separated by semicolons. For instance, to perform the three previously described actions in the same turn, the player would provide the following multi-action string: ``1 100 50; 3 -100 -10; 5 0 0''.

\subsection{State representation}
The system provides information about the player and opponent’s units. First, the game indicates the total number of units for each player’s army. Then, the system provides the following information for each one of the player and opponent's units:

\begin{itemize}
\item ID: Unique id of the unit.
\item Location: $x$, $y$ coordinates indicating the actual position of the unit on the battlefield.
\item Direction: a number indicating where the unit is looking for. It can be \texttt{Northwest} ($0$), \texttt{North} ($1$), \texttt{Northeast} ($2$), \texttt{East} ($3$), \texttt{Southeast} ($4$), \texttt{South} ($5$), \texttt{Southwest} ($6$) and \texttt{West} ($7$).
\item Life: amount of health points (See Table \ref{tab:attributes}). The unit is dead when its life reaches 0. 
\item Type: unit type for swordsmen ($0$), spearmen ($1$), cavalry ($2$) and archers ($3$).
\item Moving: Indicates if the unit is moving ($1$) or not ($0$).
\item Target: $x$, $y$ coordinates indicating where the unit is going to stop, only for friendly units (for opponent units, no target information is provided).
\end{itemize}

Therefore, the state has $1 + 9n+ 7n$ elements, where $n$ is the number of units for each player’s army. 

\subsection{Game complexity}
The number of possible actions that can be played on each turn is huge in both the draft and battle phases, due to the large battlefield size ($1920\times1080$). It also depends on the league, $1$ to $3$, selected. One possibility to handle its complexity is to artificially reduce the places where the units can be moved. According to the size of the units, we suggest defining two grids, the first one of $13\times7$ ($1920/150\approx13$,  $1080/150\approx7$) and the second one of $26\times14$ ($1920/75\approx26$,  $1080/75\approx14$). Note that the units can be always be moved to any place of the battlefield. The use of the grid is just for reducing the complexity of the game. It is suggested to be used in the first stages of the implementation of the agent, or for beginners.

\begin{table}[t]
\centering
\caption{Number of possible actions in the draft phase depending of the battlefield size.}
    \begin{tabular}{|c|c|}
    \hline
    Battlefield size & $\#$ actions \\
    \hline
    \hline
    $1920\times1080$ & $8.3E6$\\
    $26\times14$ & $1.5E3$      \\
    $13\times7$  & $3.6E2$   \\
    \hline
    \end{tabular}
\label{tab:draftactions}
\end{table}

\begin{table}[t]
\centering
\caption{Number of possible actions in the battle phase depending of the league and the battlefield size.}
    \begin{tabular}{|c|c|c|c|}
    \hline
    Battlefield size & $1^{st}$ league & $2^{nd}$ league & $3^{rd}$ league \\
    \hline
    \hline
    $1920\times1080$ & $8.3E6$ & $1.9E7$ & $6.2E7$ \\
    $26\times14$     & $1.5E3$ & $3.3E3$ & $1.1E4$ \\
    $13\times7$      & $3.6E2$ & $8.2E2$ & $2.7E3$ \\
    \hline
    \end{tabular}
\label{tab:batleactions}
\end{table}

\begin{table}[t]
\centering
\caption{Number of different army combinations that can be obtained in the draft phase depending of the league and the battlefield size.}
    \begin{tabular}{|c|c|c|c|}
    \hline
    Battlefield size & $1^{st}$ league & $2^{nd}$ league & $3^{rd}$ league \\
    \hline
    \hline
    $1920\times1080$ & $1$ & $7.5E7$ & $2.5E8$ \\
    $26\times14$     & $1$ & $1.3E4$ & $4.4E4$ \\
    $13\times7$      & $1$ & $3.3E3$ & $1.1E4$ \\
    \hline
    \end{tabular}
\label{tab:armies}
\end{table}

\begin{table}[t]
\centering
\caption{Number of possible states in the battle phase.}
    \begin{tabular}{|c|c|c|c|}
    \hline
    Battlefield size & $1^{st}$ league & $2^{nd}$ league & $3^{rd}$ league \\
    \hline
    \hline
    $1920\times1080$ & $5.8E27$ & $3.0E28$ & $3.3E29$ \\
    $26\times14$     & $3.2E16$ & $7.1E16$ & $2.4E17$ \\
    $13\times7$      & $4.9E14$ & $2.5E15$ & $2.8E16$ \\
    \hline
    \end{tabular}
\label{tab:n_states}
\end{table}

Tables~\ref{tab:draftactions} and \ref{tab:batleactions} show the number of actions in both phases with respect to the size of the battlefield in the three sizes proposed: $1920\times1080$, $26\times14$ and $13\times7$. The number of actions in the draft phase depends on the size of the battlefield ($H$, $W$) and the existing number of unit types ($t=4$). This number can be calculated as: 

\begin{equation}
    H \times W \times t
\end{equation}

For instance, when the smallest grid is used ($H=13$, $W=7$), the number of actions in the draft phase is $13*7*4=3.6E2$.

The number of actions in the battle phase depends on the size of the battlefield ($H$, $W$) and the number of units on each league ($n$). $n$ is 4, 9 and 30 in leagues 1, 2 and 3, respectively. The number of actions can be calculated as: 

\begin{equation}
    H \times W \times n
\end{equation}

For instance, when the smallest grid is used ($H=13$, $W=7$) and for the third league ($n=30$), the number of actions in the battle phase is $13*7*30=2.7E3$.

Table \ref{tab:armies} shows the number of existing army combinations that can be obtained on each league. This number depends on the size of the battlefield ($H$, $W$), the existing number of unit types ($t=4$), and the number of units on each league ($n$). In this case, the formula is:

\begin{equation}
    H \times W \times t \times n
\end{equation}

For instance, when the complete battlefield is used ($H=1920$, $W=1080$) and for the second league ($n=9$), the number of actions in the battle phase is $1920*1080*4*9=7.5E7$. Note that in the first league there is just one possible combination since there is not a draft phase and the initial configuration of the army is always the same.

Table \ref{tab:n_states} shows the number of states for the three leagues and proposed battlefield sizes. This number depends on the size of the battlefield ($H$, $W$), the number of different directions ($d=8$), the number of health points ($l$) (for simplicity, we assume in these calculations that all units have the same number of health points $l=100$, see Table \ref{tab:attributes}), the existing number of unit types ($t=4$), if the unit is moving or not ($m=2$), and the number of units on each league ($n$). The number of states can be calculated as:

\begin{equation}
(H \times W \times d \times l \times t \times m \times H \times W) \times n \times (H \times W \times d \times l \times t \times m) \times n
\end{equation}

 Note that the second term, corresponding to the opponent units, do not have a second element $H \times W$ since the final target of the opponent units is unknown.
For instance, when the complete battlefield is used ($H=1920$, $W=1080$) and for the first league ($n=4$), the number of actions in the battle phase is $(1920 \times 1080 \times 8 \times 100 \times 4 \times 2 \times 1920 \times 1080) \times 4 \times (1920 \times 1080 \times 8 \times 100 \times 4 \times 2) \times 4=5.8E27$.

\subsection{Game Art}
One of the more interesting features of \textit{CodinGame} is that it is possible to replay the game. Therefore, it is possible to study how some actions have affected the game in a particular moment of the game. 

A set of assets have been designed for a better representation of the game. The units are based on the middle age and have a cartoon style (see Figure \ref{fig:units}). 

\begin{figure*}[t]
\centering
\includegraphics[width=1.0\linewidth]{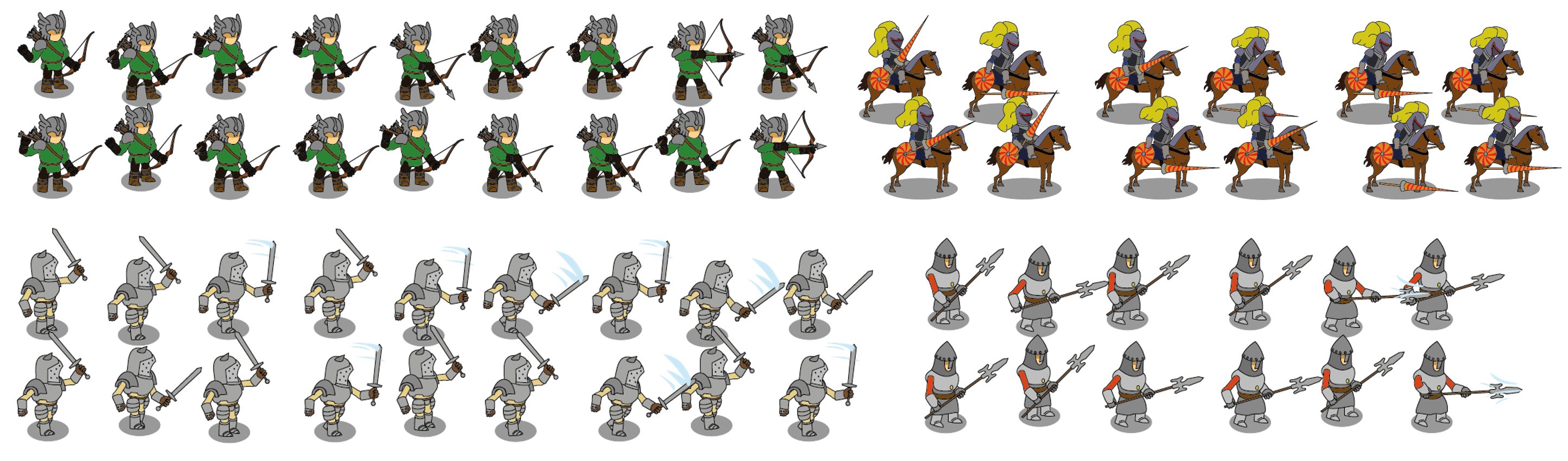}
\caption{Some animations used in the game.}
\label{fig:anim}
\end{figure*}

The game also includes animations for each state in which each unit can be found. The states are: idle, running, attacking and dead. Furthermore, there is an animation for when a unit is under arrows attack. As an example, Figure \ref{fig:anim} shows the different sprites of some of the animations used in the game. 

\section{Baseline Agents}
\label{sec:agents}
Several AI agents have been developed as baselines for the proposed game. They are briefly explained as follows:

\subsubsection{StayStatic ($SS$)} All units stand still during the battle. A predefined army is always selected in the draft. The knights are in the flanks, spearmen and swordsmen in the middle and archers behind. The units never move but they can fight when colliding with an opponent. Besides, the archers can shoot arrows to opponent troops into the attacking range.

\subsubsection{AlwaysForward ($AF$)} All units always move forward. The predefined army is the same as in $SS$.

\subsubsection{Random ($RND$)} All units select random destinations. The predefined army is the same as in $SS$.

\subsubsection{Heuristic ($\Lambda$)} It uses human knowledge in both phases. In the draft, the agent tries to pick the unit to have an advantage against the opponent. For instance, if the opponent selected in the previous turn a Knight, it will pick a Spearman. The agent has some rules to avoid to choose too many units of the same type. The agent selects the position in front of the troop that can be defeated by the selected one. 

For the battle, for each unit, a heuristic function $\lambda$ is used to estimate a value indicating how good is to attack each enemy unit. The enemy unit with the biggest value is the one selected as the target. The heuristic function $\lambda$ has been designed as the average of $5$ factors $\phi_i$ ($\phi_i \in [0,\ldots,1], \forall{i}$ and $i \in [1,\ldots,5]$):

\begin{itemize}
    \item $\phi_1$ provides higher values if the player's unit belongs to a type with advantage with respect to the opponent one, taking into account the rock-paper-scissor combat system. It can be $1.0$, $0.5$ or $0.0$ when the opponent unit type is worse, the same or better, respectively.
    \item $\phi_2$ is $1.0$ if the player's unit avoids getting into the opponent archers attacking range; $0.0$ otherwise. 
    \item $\phi_3$ benefits having more health points than the opponent unit. It can be $1.0$, $0.5$ or $0.0$ when the opponent unit has less, the same or more health points, respectively.
    \item $\phi_4$ is $1.0$ in case of a flank attack, i.e. the attacking direction is not frontal, and $0.0$ otherwise.
    \item $\phi_5$ will be higher the closer the player's unit is to the enemy's.
\end{itemize}

In the third league, a new factor $\phi_6$ is added with a value of $1.0$ if the opponent unit is a general and $0.0$ otherwise. 

\subsubsection{Online Evolutionary Planning (OEP)} This algorithm, proposed by Justesen et al.~\cite{Justensen17journal}, evolves a vector of $N$ moves to be executed by agents in multi-action games. In the original algorithm, an initial population of vectors (individuals) is generated at random to then be evolved by the algorithm, executing actions consecutively in the forward model. The state reached when all actions are executed is evaluated to obtain a fitness for the individual.

The OEP agent implemented for \textit{TotalBotWar} uses the method described for Heuristic agent for the draft phase and for seeding the initial population in the battle. Each individual contains $N$ genes, where each gene corresponds to a unit owned by the agent and their values are the IDs of the opponent's unit to attack, i.e. the number of genes $N$ is the number of units $n$ of the army in each league. For instance, in the first league ($n=4$), a genome $[2, 1, 0, 1]$ indicates that the first unit from the OEP agent will attack the opponent unit with ID $2$, the second unit will attack unit with ID $1$, and so on. A mutation rate $\rho = 0.1$ is applied to each gene to change the target to attack. The resultant states are evaluated using the same $\phi_i$ factors than in $\Lambda$, but adding a new one that rewards individuals who target the same opponent unit more than once.

\section{Experiments}
\label{sec:exps}

Several games have been played using the agents developed and described in Section~\ref{sec:agents}. Table \ref{tab:exps} shows the win-rate of the agents playing as the first player. In all cases, results are reported using the third league, the complete battlefield size (i.e. no grid is used) and $100$ games. As expected, $\Lambda$ and \textit{OEP} agents overcome the simplest baselines. Surprisingly there is a tie between $\Lambda$ and \textit{OEP} agents. This is likely due to the \textit{OEP} agent not having enough time (with the time limit constrain of \textit{CodingGame} platform) to evolve stronger action selections, being unable to find better recommendations than the ones initially provided by the $\Lambda$ agent. 

We have also tested both algorithms (\textit{OEP} vs $\Lambda$) in leagues 2 and 1, obtaining a win-rate for the \textit{OEP} of $0.65$ and $0.90$, respectively. In these cases, the game is less complex than in the case of the league 3. Therefore, the \textit{OEP} agent performs more iterations in the allowed budget time and it is able to find better move recommendations.

\begin{table}[t]
    \centering
        \caption{Win-rate of agents tested in League 3.}
    \begin{tabular}{|c|c|c|c|c|c|}
\hline
 & \textit{SS} & \textit{AF} & \textit{RND} & $\Lambda$ & \textit{OEP} \\
\hline
\hline
\textit{SS}    &     & 0.0  & 0.2  & 0.0  &  0.0  \\
\textit{AF}    & 1.0 &      & 0.5  & 0.0  &  0.05 \\
\textit{RND}   & 0.8 & 0.5  &      & 0.2  &  0.1  \\
$\Lambda$      & 1.0 & 1.0  & 0.8  &      &  0.5 \\
\textit{OEP}   & 1.0 & 0.95 & 0.90 & 0.5 &       \\
\hline
    \end{tabular}

    \label{tab:exps}
\end{table}

\section{Conclusions}
\label{sec:conclusions}
This paper presents a new game for game AI: \textit{TotalBotWar}. The game introduces interesting features and challenges for AI, as it presents a pseudo real-time decision making problem in a large and continue state and action space. It also provides an interesting challenge for drafting policies, army composition and tactical planning. The paper suggests different possibilities for discretizing the state space, and also benchmarks a state-of-the-art algorithm, Online Evolutionary Planning (OEP), which shows good results in the simpler scenarios but can't outperform domain-knowledge rule-based agents in the complex ones due to the time limitation constrain in \textit{CodingGame} platform.

Future work can span in multiple directions. Regarding the game, more complex units, rules and terrain features could be added. In terms of agents, sophisticated agents and techniques can be object of research to outperform the proposed baselines. Finally, we plan to propose this benchmark as a new competition in the future for game playing AI research.

\bibliographystyle{IEEEtran}
\bibliography{references}
\end{document}